\documentclass[sigconf]{acmart}
\AtBeginDocument{%
  \providecommand\BibTeX{{%
    \normalfont B\kern-0.5em{\scshape i\kern-0.25em b}\kern-0.8em\TeX}}}

\setcopyright{iw3c2w3}
\copyrightyear{2023}
\acmYear{2023}
\acmConference[WORKS '23]{Make sure to enter the correct
  conference title from your rights confirmation emai}{November 12,
  2023}{Denver, CO}
\acmBooktitle{WORKS '23: 18th Workshop on Workflows in Support of Large-Scale Science,
 November 12, 2023, Denver, CO} 
\begin{document}

\title{Leveraging Large Language Models to Build and Execute Computational Workflows}


%
%
\author{Alejandro Duque}
\orcid{0000-0003-4046-518X}
\affiliation{%
 \institution{Universidad San Francisco de Quito}
 \city{Quito}
 \country{Ecuador}
}
\email{aduquea@estud.usfq.edu.ec}

\author{Abdullah Syed}
\orcid{0000-0000-0000-0000}
\affiliation{%
 \institution{University of Missouri}
 \city{Columbia}
 \state{Missouri}
 \country{USA}
}
\email{aasgdc@umsystem.edu}

%\author{Eklavya Tyagi}
%\affiliation{
% \institution{University of Illinois Urbana Champaign}
% \city{Urbana}
% \state{Illinois}
% \country{USA}
%}
%\email{etyagi2@illinois.edu}

\author{Kastan V. Day}
\orcid{0000-0002-1989-3406}
\affiliation{%
  \institution{National Center for Supercomputing Applications}
  \streetaddress{1205 W Clark St}
 \city{Urbana}
 \state{Illinois}
 \country{USA}
}
\email{kvday2@illinois.edu}

\author{Matthew J. Berry}
\orcid{0000-0002-8197-9577}
\affiliation{%
  \institution{Visual Analytics\\ National Center for Supercomputing Applications}
  \streetaddress{1205 W Clark St}
  \city{Urbana}
  \state{Illinois}
  \country{USA}
  \postcode{61801}
}
\email{mjberry@illinois.edu}

\author{Daniel S. Katz}
\orcid{0000-0001-5934-7525}
\affiliation{%
  \institution{NCSA \& CS \& ECE \& iSchool\\ University of Illinois Urbana Champaign}
  \streetaddress{1205 W Clark St}
  \city{Urbana}
  \state{Illinois}
  \country{USA}
  \postcode{61801}
}
\email{d.katz@ieee.org}

\author{Volodymyr V. Kindratenko}
\orcid{0000-0002-9336-4756}
\affiliation{%
  \institution{Center for AI Innovation\\ National Center for Supercomputing Applications}
  \streetaddress{1205 W Clark St}
  \city{Urbana}
  \state{Illinois}
  \country{USA}
  \postcode{61801}
}
\email{kindrtnk@illinois.edu}

%\affiliation{%
%  \institution{Inria Paris-Rocquencourt}

\renewcommand{\shortauthors}{Duque and Syed, et al.}

\begin{abstract}
  The recent development of large language models (LLMs) with multi-billion parameters, coupled with the creation of user-friendly application programming interfaces (APIs), has paved the way for automatically generating and executing code in response to straightforward human queries. This paper explores how these emerging capabilities can be harnessed to facilitate complex scientific workflows, eliminating the need for traditional coding methods. We present initial findings from our attempt to integrate Phyloflow with OpenAI's function-calling API, and outline a strategy for developing a comprehensive workflow management system based on these concepts.
\end{abstract}





\received{16 August 2023}
\received[revised]{27 September 2023}
%\received[accepted]{TBD}

\maketitle

\section{Introduction}

Scientific computing workflows are important tools in contemporary research paradigms that involve data processing and computation. Many approaches have been developed over the past decade to describe the computational work to be carried out, e.g., the Workflow Description Language (WDL)~\cite{WDL}, Common Workflow Language (CWL) \cite{CWL}, as well as ways to actually carry them out, e.g., Cromwell~\cite{cromwell}, Toil~\cite{toil}, Apache Airflow~\cite{airflow}. Some of these description languages and tools are domain-specific while others are generic and are used across multiple domains. In all cases, however, the researchers need to learn how to describe the workflows for their specific applications as well as how to use the workflow management framework on their specific computer systems.

In this position paper, we argue that in the near future it will be possible to construct and utilize integrated scientific workflow description and execution systems using Natural Language within a chatbot-like environment, e.g., within ChatGPT's framework. We observe that even today ChatGPT is capable of generating codes in various workflow description languages which then can be executed with the help of emerging plugins and new features, such LangChain~\cite{langchain}, function calling~\cite{oaifc}, Toolformer~\cite{toolformer}, code interpreter~\cite{oaici}, to name a few. We believe that a next-generation workflow management system will simply provide a human-language-based interface to describe the work to be carried out, monitor its progression, and present the results, and will call underlying tools and methods to carry out a complex chain of computations in response.  This will be a game-changing capability that will greatly simplify the process of applying complex computational pipelines by domain experts without any coding experience.

The paper is structured as follows: In \S\ref{sec:review}, we review existing Large Language Model (LLM) capabilities for code generation and emerging techniques for code execution within the ChatGPT framework; in \S\ref{sec:PrelimRes} we provide initial results with function calling API; and in \S\ref{sec:NextGen} we put forward a proposal for a ChatGPT-based workflow management system. 

\section{Review of Emerging Technologies}\label{sec:review}

\subsection{Code Generation} \label{CodeGen}

Modern LLM frameworks, such as ChatGPT~\cite{oaichat}, possess advanced code generation capabilities. These models have been trained on a diverse range of internet text, including programming code from GitHub, coding questions and answers from Stack Overflow, and various documentation, which enables them to generate code snippets in response to user prompts. As the result, tools such as ChatGPT can assist with many coding tasks in many languages, can support code analysis and debugging, and can even write complete programs~\cite{buscemi2023comparative}. GitHub Copilot~\cite{copilot} is one such example of a widely-used tool. However, it is important to note that while such tools can generate syntactically correct code, they do not inherently understand the code's semantics~\cite{chen2021evaluating}. In practice this means they cannot guarantee the functional correctness of the generated code and it is up to the user to ensure that the produced code will accomplish the required task. Nevertheless, their ability to generate code provides a valuable tool for developers, aiding in tasks ranging from code completion to generating boilerplate code.

\subsection{Code Execution}

Code execution is not something ChatGPT or other LLMs are directly capable of. However, code execution has gained a lot of interest and is a very active area of research and development. It can be achieved in a number of ways using several emerging methods; here we briefly touch on some of them, particularly with regards to those related to ChatGPT framework. 

Using the LangChain library~\cite{langchain}, one can connect (chain) multiple components to create applications that utilize LLMs. For code execution, we can use \textit{agents} through which an LLM can write and execute Python code. The \textit{agents} call user-supplied \textit{tools}: functions written to interact with the world. An extensive collection of such \textit{tools} already exists~\cite{lctools}. The interface to \textit{tools} is plain text; however, even this simple interface is sufficient to accomplish rather complex tasks. Here a language model itself is used as a reasoning engine to determine which actions to take and in which order. Thus, an LLM-generated code can be executed with LangChain by creating an \textit{agent} that can call an appropriate \textit{tool} to compile (if needed) and then execute the code.

Unlike LangChain in which the programmer still needs to call agents, with recently introduced function calling API~\cite{oaifc} one can simply describe functions to an LLM and have the model generate a JSON object containing arguments to call those functions. Thus, based on a user-supplied description of a collection of functions, the LLM itself can infer which function to call, produce the necessary interface for the call, call the function, and process the output. 

Toolformer~\cite{toolformer} is another approach developed to make an LLM to execute a function. Toolformer is a ``model trained to decide which APIs to call, when to call them, what arguments to pass, and how to best incorporate the results into future token prediction.'' The training is achieved in a self-supervised manner, needing just a few examples for each API call, and potentially using already existing API documentation for an LLM to learn from.

Code interpreter~\cite{oaici}, a.k.a. Advanced Data Analytics, is a ChatGPT plugin that is able to run a Python code in a sandbox environment to make sure it works. This fixes the original problem with the code generation mentioned in \S\ref{CodeGen}. Thus, with this plugin one can ask ChatGPT to write a code and immediately execute it. 

It is important to note that we are still in early days of code execution and new projects and products emerge quite regularly.

\section{Preliminary Results} \label{sec:PrelimRes}

As a demonstration, we have used Phyloflow~\cite{6999303} as an existing workflow example and investigated how OpenAI's function calling API can be used to streamline the creation and execution of different tasks. The code for the work presented here is available in GitHub~\cite{ourwork}. Phyloflow is a tool for performing phylogenetic tree computations that was initially developed in WDL and later ported to Python using the Parsl library~\cite{parsl}. The Parsl implementation consisted of a Parsl app for each task that corresponds to a data processing step within Phyloflow. A  workflow is created by connecting Parsl apps through inter-application dependencies. It is important to note that for the first step of this Parsl workflow, we use physical files, and from there we work on data futures, which are promises of a file that will be generated by another running instance of a Parsl app (AppFuture).

Phyloflow employs WDL to perform calculations for phylogenetic analysis based on input data. This analysis involves computing clusters which are then used to construct phylogeny trees that can be visualized. This process is executed in multiple steps, beginning with the task `vcf-transform.' VCF-transform receives a VCF (Variant Calling Format) file and extracts data from it, and transforms the data into the input format of `pyclone-vi', which allows for mutation clustering calculations to be performed. Next, a TSV file containing the mutagenic data is created. The next task, 'pyclone-vi', processes the previously generated TSV file, performing the mutation clustering calculations. This yields clusters of mutations that share evolutionary relationships. The workflow subsequently reformats this data file for processing with SPRUCE (Somatic Phylogeny Reconstruction using Combinatorial Enumeration) in a separate workflow step. SPRUCE is an algorithm utilized for inferring phylogenies that describe the evolutionary history of a tumor. The final task, `spruce-phylogeny', takes the SPRUCE-formatted TSV file as an input, and generates a JSON file that contains the computed information necessary to visualize the evolution of a tumor. To illustrate this, the first workflow step is shown:

{\small
\begin{verbatim}
task vcf_transform {
    input {
        File vcf_file
        String vcf_type #only 'mutect' currently supported
    }
    command {
        mkdir pyclone_samples
        out_dir=$(pwd)
        cd /code
        sh vcf_transform_entrypoint.sh ${vcf_type} ${vcf_file} \
            $out_dir/headers.json \
            $out_dir/mutations.json \
            $out_dir/pyclone_vi_formatted.tsv \
            $out_dir/pyclone_samples/
    }
    output {
        File response = stdout()
        File err_response = stderr()
        File headers_json = 'headers.json'
        File mutations_json = 'mutations.json'
        Array[File] pyclone_formatted_tsvs =
          glob("pyclone_samples/*.tsv")
    }
}

\end{verbatim}
}

To enable the OpenAI function call API with Phyloflow, we created several functions that serve as adapters for Parsl apps. For each Parsl app, we created a \textit{function\_call\_from\_file}, which receives the paths to the physical files, and a \textit{function\_call\_from\_futures}, which receives the identifiers of the AppFutures on which the Parsl app depends. The difference between the two is that a \textit{function\_call\_from\_futures} first retrieves the AppFutures of the received IDs, and their DataFutures are extracted to be used as inputs. From there, the operation is identical: generate a new ID, generate a directory for outputs, run the ParslApp, index the AppFuture reference along with its ID in a global access dictionary and return the ID. This ID binding scheme with AppFutures was required to communicate with the OpenAI API.

Following the OpenAI specifications, we wrote function descriptions in JSON format for all of the \textit{function\_call\_from\_files} and \textit{function\_call\_from\_futures}. Here are some examples of the function descriptions:

{\small
\begin{verbatim}
functions = [
    {
        'name': 'fcall_pyclone_vi_from_files',
        'description': 'Computes mutation clusters from 
                        vcf_transformed file',
        'parameters': {
            'type': 'object',
            'properties': {
                'pyclone_vi_formatted': {
                    'type': 'string',
                    'description': 'The path to the 
                    pyclone_vi_formatted file outputed 
                    by the vcf_transform'
                },
            },
            'required': ['pyclone_vi_formatted']
        }
    },
    {
        'name': 'fcall_pyclone_vi_from_futures',
        'description': 'Computes mutation clusters from 
                        a vcf_transform AppFuture id',
        'parameters': {
            'type': 'object',
            'properties': {
                'vcf_future_id': {
                    'type': 'string',
                    'description': 'The vcf_transform id'
                },
            },
            'required': ['vcf_future_id']
        }
    }
]

\end{verbatim}
}

The communication scheme with the OpenAI API consists of sending this set of descriptions together with a natural language instruction prompted by the user. The job of the LLM is to determine which function needs to be executed to fulfill the statement, as well as the parameters to send to the function. By doing this, we were able to run individual Parsl applications within the workflow. However, what we really need is to chain the execution of several Parsl apps to generate complete workflow executions. This is where the notion of adding context and making successive API calls comes into play.

A predefined context is added, just like any other user message, that helps to better interpret any instruction. With this context and the user's message, the request to the API is made. The API responds with its choice of function to call. The function is executed, immediately returning the ID linked to the AppFuture. For the next API request, two new messages are added. The first message partially includes the previous response from the API, specifically the section of the message with the choice of the function to call is used. The second message is a new user message indicating the ID assigned to the newly executed Parsl app. By adding both messages, the AI understands which step it is in relative to the user's instructions and can also execute subsequent steps by having access to the scheduled AppFuture ID. This process is repeated until the stop flag is found in the API response. Below is the execution of an example user instruction:

{\small
\begin{verbatim}

Context:
    If you are asked to execute one single task receive file names
   If you are asked to execute multiple tasks:
        Receive file names for the first task
        Send the future ids to the other tasks

User: 
Help me with two things: 
First: transform the vcf file 
./example_data/VEP_raw.A25.mutect2.filtered.snp.vcf.
Second: execute pyclone-vi on the file outputed in the first step.

Function Calling
Function Name:  fcall_vcf_transform_from_files
Function Args:  
{'vep_vcf': './example_data/VEP_raw.A25.mutect2.filtered.snp.vcf'}
<AppFuture at 0x7f90af178b90 state=pending>

Task scheduled with AppFuture id: future_5_run_vcf_transform

Function Calling
Function Name:  fcall_pyclone_vi_from_futures
Function Args:  {'vcf_future_id': 'future_5_run_vcf_transform'}
<AppFuture at 0x7f9072014490 state=pending>

Task scheduled with AppFuture id: future_6_run_pyclone_vi

DONE

\end{verbatim}
}

Although the use of function calling has shown promise for executing workflows, our current implementation has at least two clear limitations. The first is that exceptions are not handled at the moment, which means that if the API executes a wrong function call, the program cannot recover from the failure. Optimally, the error should be forwarded to the API so that it can propose alternatives. The second limitation is that composing more complex workflows will eventually hit the token limit, for which there is no straightforward solution in the proposed scheme; we would need to invent a hierarchical schema for task decomposition. 

\section{Proposal for Next-Gen Workflow Engine} \label{sec:NextGen}

The system we discuss in \S\ref{sec:PrelimRes} consists of two primary components: the execution of OpenAI API calls on OpenAI's servers and the processing stages of the Phyloflow application on our own servers. The current prototype sequentially executes these steps without much consideration for the results produced at each step. However, a more advanced workflow engine should ensure two things: 

1. The current step is executed as expected, free of errors or warnings, and produces the anticipated outcome.

2. The next step in the sequence can be executed given the outcome of the current step, the available computational resources, and other constraints.

We envision a workflow engine that accepts a high-level \textit{description} of the work, provided in natural language. This description is then translated into multiple steps (a \textit{plan}) based on available functional units such as executables or API calls. The engine then attempts to execute each step from the \textit{plan}, taking into account hardware constraints such as the type of compute servers, available memory, storage size, while ensuring the task's completeness and correctness. If a task fails or the outcome is not as expected, the plan execution engine invokes a \textit{debugger}. The \textit{debugger's} role is to identify the issue so the task can be re-executed or the \textit{plan} can be modified if necessary.

\begin{figure}[h]
  \centering
  \includegraphics[width=\linewidth]{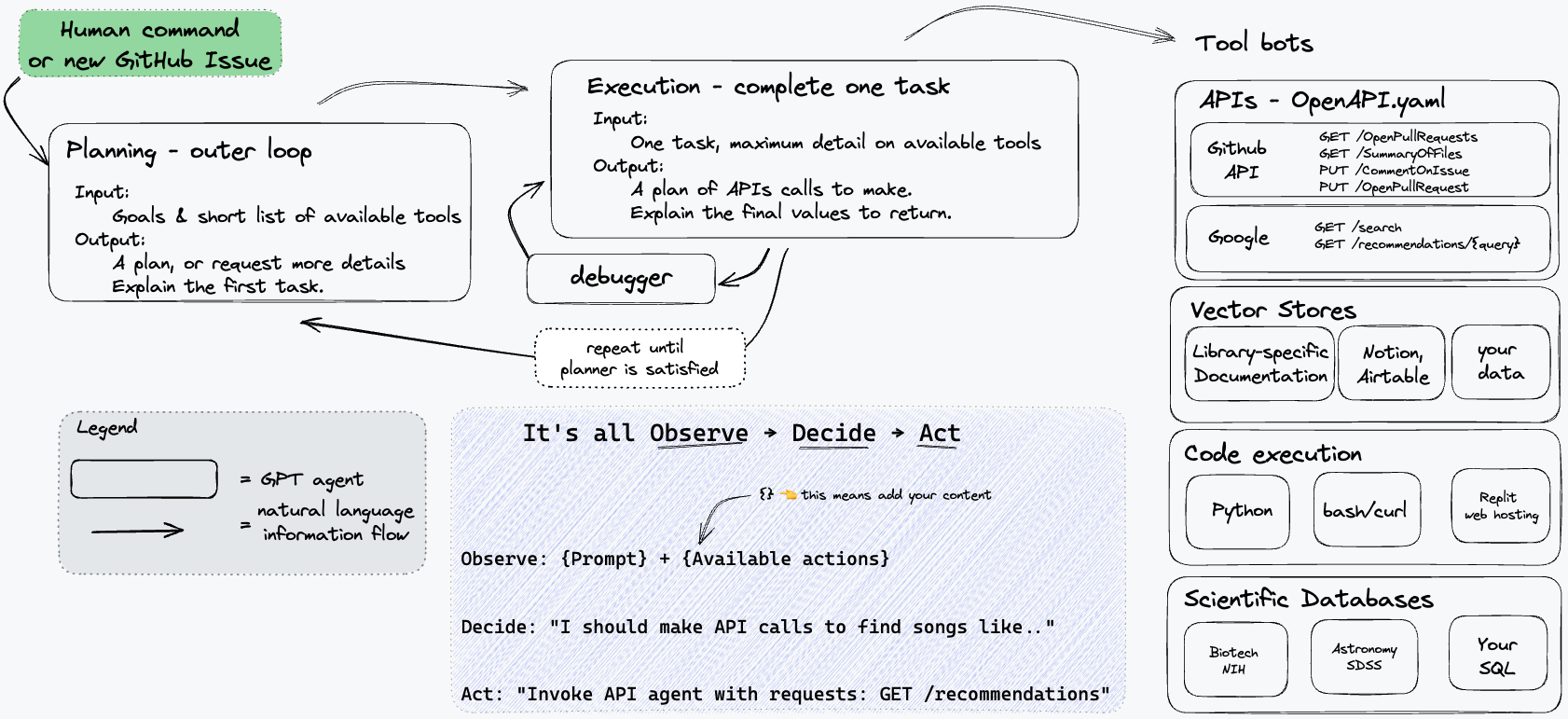}
  \caption{LLM agents collaborating to execute the workflow.}
  \label{fig1}
\end{figure}

Figure \ref{fig1} illustrates the overall structure of this approach. The \textit{planner}, \textit{executor}, and \textit{debugger} are all AI agents that use LLM to process textual input, either to execute a task or to analyze and validate the execution results. A \textit{human} operator may also be involved if the \textit{debugger} cannot resolve the issue, or if there's a need to resolve ambiguities and make decisions.

\section{Conclusions and Future Work}

In \S\ref{sec:PrelimRes}, we successfully showcased a working prototype of a workflow engine that can execute a variety of tasks based on straightforward high-level user instructions. This accomplishment was made possible by leveraging OpenAI's function-calling API. This API infers tasks from the user's textual description and translates them into user-defined functions, which constitute the computational backbone of the actual scientific computing workflow. Furthermore, in  \S\ref{sec:NextGen}, we outlined our ambitious vision for an advanced workflow execution engine. This engine is designed to manage complex, multi-stage workflows across an extensive computing infrastructure, all under the control of a natural human language interface. We are actively progressing towards making this vision a reality.

\begin{acks}
This work was partially supported by NSF award 2050195.
\end{acks}

\bibliographystyle{ACM-Reference-Format}
\bibliography{sample-base}


\begin{thebibliography}{17}

%%% the Web address.

\ifx \showCODEN    \undefined \def \showCODEN     #1{\unskip}     \fi
\ifx \showDOI      \undefined \def \showDOI       #1{#1}\fi
\ifx \showISBNx    \undefined \def \showISBNx     #1{\unskip}     \fi
\ifx \showISBNxiii \undefined \def \showISBNxiii  #1{\unskip}     \fi
\ifx \showISSN     \undefined \def \showISSN      #1{\unskip}     \fi
\ifx \showLCCN     \undefined \def \showLCCN      #1{\unskip}     \fi
\ifx \shownote     \undefined \def \shownote      #1{#1}          \fi
\ifx \showarticletitle \undefined \def \showarticletitle #1{#1}   \fi
\ifx \showURL      \undefined \def \showURL       {\relax}        \fi
% invisible to TeX
\providecommand\bibfield[2]{#2}
\providecommand\bibinfo[2]{#2}
\providecommand\natexlab[1]{#1}
\providecommand\showeprint[2][]{arXiv:#2}

\bibitem[lan(2023)]%
        {langchain}
 \bibinfo{year}{2023}\natexlab{}.
\newblock \bibinfo{title}{LangChain}.
\newblock
  \bibinfo{howpublished}{\url{https://github.com/langchain-ai/langchain}}.
\newblock
\newblock
\shownote{Accessed: 2023-08-16}.


\bibitem[WDL(2023)]%
        {WDL}
 \bibinfo{year}{2023}\natexlab{}.
\newblock \bibinfo{title}{Workflow Description Language (WDL)}.
\newblock \bibinfo{howpublished}{\url{https://github.com/openwdl/wdl}}.
\newblock
\newblock
\shownote{Accessed: 2023-08-16}.


\bibitem[Alvarez-Jarreta et~al\mbox{.}(2014)]%
        {6999303}
\bibfield{author}{\bibinfo{person}{J. Alvarez-Jarreta}, \bibinfo{person}{G. de
  Miguel~Casado}, {and} \bibinfo{person}{E. Mayordomo}.}
  \bibinfo{year}{2014}\natexlab{}.
\newblock \showarticletitle{PhyloFlow: A fully customizable and automatic
  workflow for phylogenetic reconstruction}. In \bibinfo{booktitle}{\emph{2014
  IEEE International Conference on Bioinformatics and Biomedicine (BIBM)}}.
  \bibinfo{pages}{1--7}.
\newblock
\urldef\tempurl%
\url{https://doi.org/10.1109/BIBM.2014.6999303}
\showDOI{\tempurl}


\bibitem[Apache({[n.\,d.]})]%
        {airflow}
\bibfield{author}{\bibinfo{person}{Apache}.}
  \bibinfo{year}{[n.\,d.]}\natexlab{}.
\newblock \bibinfo{title}{Airflow}.
\newblock \bibinfo{howpublished}{\url{https://airflow.apache.org}}.
\newblock


\bibitem[Babuji et~al\mbox{.}(2019)]%
        {parsl}
\bibfield{author}{\bibinfo{person}{Y. Babuji}, \bibinfo{person}{A. Woodard},
  \bibinfo{person}{Z. Li}, \bibinfo{person}{D. Katz}, \bibinfo{person}{B.
  Clifford}, \bibinfo{person}{R.~Kumar L.}, \bibinfo{person}{Lacinski},
  \bibinfo{person}{R. Chard}, \bibinfo{person}{J. Wozniak}, \bibinfo{person}{I.
  Foster}, \bibinfo{person}{M. Wilde}, {and} \bibinfo{person}{K. Chard}.}
  \bibinfo{year}{2019}\natexlab{}.
\newblock \showarticletitle{Parsl: Pervasive Parallel Programming in Python}.
  In \bibinfo{booktitle}{\emph{Proceedings of the 28th International Symposium
  on High-Performance Parallel and Distributed Computing}} (Phoenix, AZ, USA)
  \emph{(\bibinfo{series}{HPDC '19})}. \bibinfo{publisher}{Association for
  Computing Machinery}, \bibinfo{address}{New York, NY, USA},
  \bibinfo{pages}{25–36}.
\newblock
\showISBNx{9781450366700}
\urldef\tempurl%
\url{https://doi.org/10.1145/3307681.3325400}
\showDOI{\tempurl}


\bibitem[Buscemi(2023)]%
        {buscemi2023comparative}
\bibfield{author}{\bibinfo{person}{A. Buscemi}.}
  \bibinfo{year}{2023}\natexlab{}.
\newblock \bibinfo{title}{A Comparative Study of Code Generation using ChatGPT
  3.5 across 10 Programming Languages}.
\newblock
\newblock
\showeprint[arxiv]{2308.04477}~[cs.SE]


\bibitem[Chase(2023)]%
        {lctools}
\bibfield{author}{\bibinfo{person}{H. Chase}.} \bibinfo{year}{2023}\natexlab{}.
\newblock \bibinfo{title}{langchain API Reference}.
\newblock
  \bibinfo{howpublished}{\url{https://api.python.langchain.com/en/latest/api_reference.html##module-langchain.tools}}.
\newblock
\newblock
\shownote{Accessed: 2023-08-16}.


\bibitem[Chen et~al\mbox{.}(2021)]%
        {chen2021evaluating}
\bibfield{author}{\bibinfo{person}{M. Chen}, \bibinfo{person}{J. Tworek},
  \bibinfo{person}{H. Jun}, \bibinfo{person}{Q. Yuan}, \bibinfo{person}{H.
  Pinto}, \bibinfo{person}{J. Kaplan}, \bibinfo{person}{H. Edwards},
  \bibinfo{person}{Y. Burda}, \bibinfo{person}{N. Joseph}, \bibinfo{person}{G.
  Brockman}, \bibinfo{person}{A. Ray}, \bibinfo{person}{R. Puri},
  \bibinfo{person}{G. Krueger}, \bibinfo{person}{M. Petrov},
  \bibinfo{person}{H. Khlaaf}, \bibinfo{person}{G. Sastry}, \bibinfo{person}{P.
  Mishkin}, \bibinfo{person}{B. Chan}, \bibinfo{person}{S. Gray},
  \bibinfo{person}{N. Ryder}, \bibinfo{person}{M. Pavlov}, \bibinfo{person}{A.
  Power}, \bibinfo{person}{L. Kaiser}, \bibinfo{person}{M. Bavarian},
  \bibinfo{person}{C. Winter}, \bibinfo{person}{P. Tillet}, \bibinfo{person}{F.
  Such}, \bibinfo{person}{D. Cummings}, \bibinfo{person}{M. Plappert},
  \bibinfo{person}{F. Chantzis}, \bibinfo{person}{E. Barnes},
  \bibinfo{person}{A. Herbert-Voss}, \bibinfo{person}{W. Guss},
  \bibinfo{person}{A. Nichol}, \bibinfo{person}{A. Paino}, \bibinfo{person}{N.
  Tezak}, \bibinfo{person}{J. Tang}, \bibinfo{person}{I. Babuschkin},
  \bibinfo{person}{S. Balaji}, \bibinfo{person}{S. Jain}, \bibinfo{person}{W.
  Saunders}, \bibinfo{person}{C. Hesse}, \bibinfo{person}{A. Carr},
  \bibinfo{person}{J. Leike}, \bibinfo{person}{J. Achiam}, \bibinfo{person}{V.
  Misra}, \bibinfo{person}{E. Morikawa}, \bibinfo{person}{A. Radford},
  \bibinfo{person}{M. Knight}, \bibinfo{person}{M. Brundage},
  \bibinfo{person}{M. Murati}, \bibinfo{person}{K. Mayer}, \bibinfo{person}{P.
  Welinder}, \bibinfo{person}{B. McGrew}, \bibinfo{person}{D. Amodei},
  \bibinfo{person}{S. McCandlish}, \bibinfo{person}{I. Sutskever}, {and}
  \bibinfo{person}{W. Zaremba}.} \bibinfo{year}{2021}\natexlab{}.
\newblock \bibinfo{title}{Evaluating Large Language Models Trained on Code}.
\newblock
\newblock
\showeprint[arxiv]{2107.03374}~[cs.LG]


\bibitem[Crusoe et~al\mbox{.}(2022)]%
        {CWL}
\bibfield{author}{\bibinfo{person}{M. Crusoe}, \bibinfo{person}{S. Abeln},
  \bibinfo{person}{A. Iosup}, \bibinfo{person}{P. Amstutz}, \bibinfo{person}{J.
  Chilton}, \bibinfo{person}{N. Tijani\'{c}}, \bibinfo{person}{H. M\'{e}nager},
  \bibinfo{person}{S. Soiland-Reyes}, \bibinfo{person}{B. Gavrilovi\'{c}},
  \bibinfo{person}{C. Goble}, {and} \bibinfo{person}{The~CWL Community}.}
  \bibinfo{year}{2022}\natexlab{}.
\newblock \showarticletitle{Methods Included: Standardizing Computational Reuse
  and Portability with the Common Workflow Language}.
\newblock \bibinfo{journal}{\emph{Commun. ACM}} \bibinfo{volume}{65},
  \bibinfo{number}{6} (\bibinfo{date}{May} \bibinfo{year}{2022}),
  \bibinfo{pages}{54–63}.
\newblock
\showISSN{0001-0782}
\urldef\tempurl%
\url{https://doi.org/10.1145/3486897}
\showDOI{\tempurl}


\bibitem[Duque and Syed(2023)]%
        {ourwork}
\bibfield{author}{\bibinfo{person}{A. Duque} {and} \bibinfo{person}{A. Syed}.}
  \bibinfo{year}{2023}\natexlab{}.
\newblock \bibinfo{title}{Phyloflow-Parsl Implementation}.
\newblock
  \bibinfo{howpublished}{\url{https://github.com/grimloc-aduque/Phyloflow-Parsl-Implementation}}.
\newblock
\newblock
\shownote{Accessed: 2023-08-16}.


\bibitem[GitHub(2022)]%
        {copilot}
\bibfield{author}{\bibinfo{person}{GitHub}.} \bibinfo{year}{2022}\natexlab{}.
\newblock \bibinfo{title}{Github Copilot}.
\newblock \bibinfo{howpublished}{\url{https://docs.github.com/en/copilot}}.
\newblock
\newblock
\shownote{Accessed: 2023-08-16}.


\bibitem[Institute(2023)]%
        {cromwell}
\bibfield{author}{\bibinfo{person}{Broad Institute}.}
  \bibinfo{year}{2023}\natexlab{}.
\newblock \bibinfo{title}{Cromwell}.
\newblock
  \bibinfo{howpublished}{\url{https://github.com/broadinstitute/cromwell}}.
\newblock
\newblock
\shownote{Accessed: 2023-08-16}.


\bibitem[OpenAI(2022)]%
        {oaichat}
\bibfield{author}{\bibinfo{person}{OpenAI}.} \bibinfo{year}{2022}\natexlab{}.
\newblock \bibinfo{title}{ChatGPT}.
\newblock \bibinfo{howpublished}{\url{https://chat.openai.com/chat}}.
\newblock
\newblock
\shownote{Accessed: 2023-08-16}.


\bibitem[OpenAI(2023a)]%
        {oaici}
\bibfield{author}{\bibinfo{person}{OpenAI}.} \bibinfo{year}{2023}\natexlab{a}.
\newblock \bibinfo{title}{ChatGPT plugins: Code Interpreter}.
\newblock
  \bibinfo{howpublished}{\url{https://openai.com/blog/chatgpt-plugins##code-interpreter}}.
\newblock
\newblock
\shownote{Accessed: 2023-08-16}.


\bibitem[OpenAI(2023b)]%
        {oaifc}
\bibfield{author}{\bibinfo{person}{OpenAI}.} \bibinfo{year}{2023}\natexlab{b}.
\newblock \bibinfo{title}{GPT Models: Function Calling}.
\newblock
  \bibinfo{howpublished}{\url{https://platform.openai.com/docs/guides/gpt/function-calling}}.
\newblock
\newblock
\shownote{Accessed: 2023-08-16}.


\bibitem[Schick et~al\mbox{.}(2023)]%
        {toolformer}
\bibfield{author}{\bibinfo{person}{T. Schick}, \bibinfo{person}{J. Dwivedi-Yu},
  \bibinfo{person}{R. Dessì}, \bibinfo{person}{R. Raileanu},
  \bibinfo{person}{M. Lomeli}, \bibinfo{person}{L. Zettlemoyer},
  \bibinfo{person}{N. Cancedda}, {and} \bibinfo{person}{T. Scialom}.}
  \bibinfo{year}{2023}\natexlab{}.
\newblock \bibinfo{title}{Toolformer: Language Models Can Teach Themselves to
  Use Tools}.
\newblock
\newblock
\showeprint[arxiv]{2302.04761}~[cs.CL]


\bibitem[Vivian et~al\mbox{.}(2017)]%
        {toil}
\bibfield{author}{\bibinfo{person}{J. Vivian}, \bibinfo{person}{A. Rao},
  \bibinfo{person}{F. Nothaft}, \bibinfo{person}{C. Ketchum},
  \bibinfo{person}{J. Armstrong}, \bibinfo{person}{A. Novak},
  \bibinfo{person}{J. Pfeil}, \bibinfo{person}{J. Narkizian},
  \bibinfo{person}{A. Deran}, \bibinfo{person}{A. Musselman-Brown},
  \bibinfo{person}{H. Schmidt}, \bibinfo{person}{P. Amstutz},
  \bibinfo{person}{B. Craft}, \bibinfo{person}{M. Goldman}, \bibinfo{person}{K.
  Rosenbloom}, \bibinfo{person}{M. Cline}, \bibinfo{person}{B. O’Connor},
  \bibinfo{person}{M. Hanna}, \bibinfo{person}{C. Birger}, \bibinfo{person}{W.
  Kent}, \bibinfo{person}{D. Patterson}, \bibinfo{person}{A. Joseph},
  \bibinfo{person}{J.~Zhuand~S. Zaranek}, \bibinfo{person}{G. Getz},
  \bibinfo{person}{D. Haussler}, {and} \bibinfo{person}{B. Paten}.}
  \bibinfo{year}{2017}\natexlab{}.
\newblock \bibinfo{title}{Toil enables reproducible, open source, big
  biomedical data analyses}.
\newblock , \bibinfo{numpages}{314–316}~pages.
\newblock
\urldef\tempurl%
\url{https://doi.org/10.1038/nbt.3772}
\showDOI{\tempurl}


\end{thebibliography}

\end{document}